\documentclass[10pt,twocolumn,letterpaper]{article}
\usepackage[pagenumbers]{cvpr}

\usepackage{graphicx}
\usepackage{booktabs}
\usepackage{multirow}
\usepackage{listings}
\newcommand{\sys}{BPS}

\definecolor{cvprblue}{rgb}{0.21,0.49,0.74}
\usepackage[pagebackref,breaklinks,colorlinks,allcolors=cvprblue]{hyperref}

\title{Semantics First, Pixels Later: Mission-Driven Event Streams\\for Drone Video over Bandwidth-Starved Links}
\author{Cong Xu\\Independent Researcher}

\begin{document}
\twocolumn[{%
\renewcommand\twocolumn[1][]{#1}%
\maketitle
\begin{center}
\vspace{-1.5em}
  \begin{minipage}[b]{0.325\linewidth}\centering
    \includegraphics[width=\linewidth]{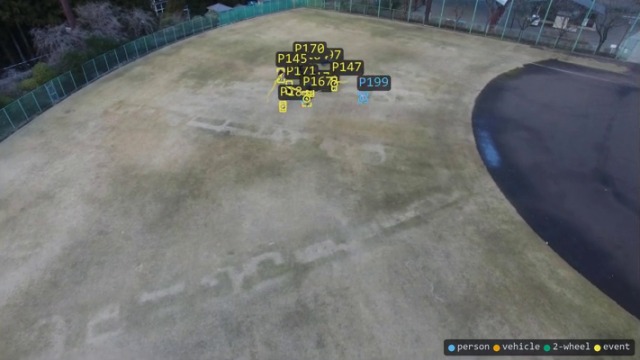}\\[1pt]
    {\small (a) Onboard view, never transmitted.}
  \end{minipage}\hfill
  \begin{minipage}[b]{0.325\linewidth}\centering
    \includegraphics[width=\linewidth]{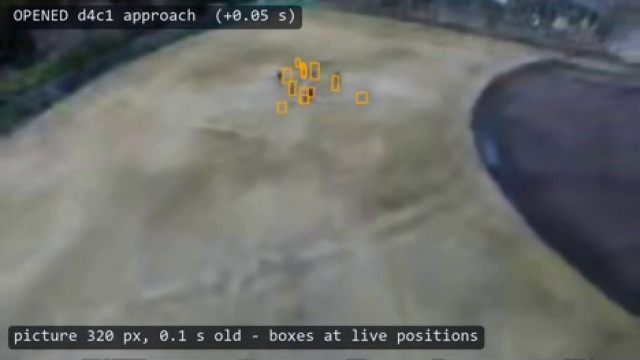}\\[1pt]
    {\small (b) \sys{} on the ground, 10\,kbps link.}
  \end{minipage}\hfill
  \begin{minipage}[b]{0.325\linewidth}\centering
    \includegraphics[width=\linewidth]{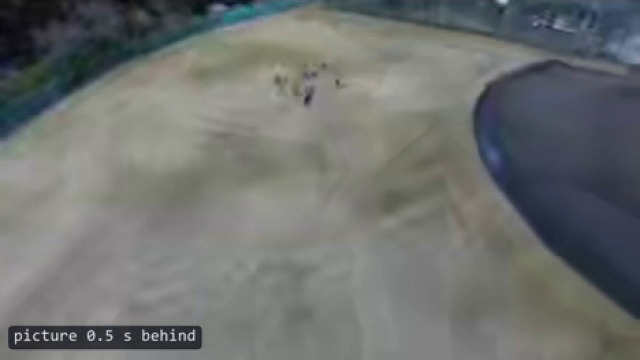}\\[1pt]
    {\small (c) Low-latency H.265, same 10\,kbps link.}
  \end{minipage}
  \captionof{figure}{The same moment (moving camera, Okutama-Action) on the ground over a 10\,kbps link. (b) \sys{}, our behavior-primitive stream: an event record (five people converging) arrived 0.05\,s after the drone decided it. The picture and the boxes are an optional layer added later and are not what the reader reads (\cref{sec:results}). (c) H.265 with low-latency settings, alone on the same link: a frame 0.5\,s old. Both are timely. A language model reading (b)'s event records answers 74\% of our development questions; its vision fine-tune reading (c)'s frames answers 49.8\%, less than the 57.4\% of a reader that receives nothing. Frames are rendered by our link emulator from the actual packet bytes.}
  \label{fig:teaser}
\end{center}
\vspace{0.4em}
}]

\begin{abstract}
A surveillance drone on a bandwidth-starved link cannot send the video its operator would need, yet the operator's questions are often not about pixels: \emph{did three people converge, when, and where?} We study what should cross a link of 1\,Mbps down to 1\,kbps instead, and in what order. \sys{}, a behavior-primitive stream, sends semantics first and pixels later: it runs a detector and a tracker onboard and sends short records (25 bytes on average) of multi-frame events selected by a mission card of 16--125 bytes, ahead of positions and pictures. An event is opened as soon as it can be decided and closed when it ends. We measure it by what a ground-side language model answers from it, under a preregistered protocol with a held-out set, against H.265 read by the vision fine-tune of the same model, H.265 on which the ground runs our own perception, a stream of quantized tracks, and onboard captions. On the development set (84\% of questions about convergence, mostly one location) the event stream answers 74.3\% of questions at 0.13\,kbps. A reader that receives nothing answers 57.4\%, and the video reader 59.6\% at 100\,kbps and 64.7\% on the source frames. When the ground instead runs our perception on the video, video matches the stream at 100\,kbps, about 800 times the stream's rate, and falls behind at 64\,kbps and below. A track stream matches it at 96 times the bytes and loses 0.03 at 17 times. On held-out data the result is weaker: with events taken from annotated tracks, an oracle, the stream beats H.265 at 100\,kbps ($+0.13$), but with onboard perception it does not ($+0.01$ [$-0.04$, $+0.06$]) and is no better than receiving nothing. Sent alone through emulated links the event records answer 0.739 at 1\,Mbps and still 0.724 at 0.3\,kbps; an every-frame track stream holds down to 1\,kbps and video with ground-side perception down to about 100\,kbps. On a Jetson Orin Nano the unchanged pipeline runs at 4.5\,fps and misses most events.
\end{abstract}

\section{Introduction}
\label{sec:intro}

Small unmanned aircraft carry cameras whose output the ground often cannot receive: tactical radios, satellite back-haul and contested spectrum deliver from about a megabit per second down to a few kilobits. A codec with low-latency settings can still push a small picture through ten kilobits per second in time (\cref{fig:teaser}), but a small picture is not an answer. The operator often asks about behavior over time: whether a vehicle stopped and stayed, whether people converged on one spot, when a quiet area came alive. This paper asks what should be sent instead of video when the answer, not the picture, is what the ground needs.

Our answer, \sys{} (behavior-primitive stream), moves the perception onboard and sends its conclusions. A detector and a tracker run on the aircraft, a streaming engine turns tracks into multi-frame behavior primitives, and a mission card tells it which primitives, classes and regions matter. A scheduler sends event records of a few tens of bytes ahead of everything else, so the stream degrades gracefully: a narrowing link costs the ground its pictures first, then positions, and the events last. An event is \emph{opened} as soon as the engine can decide it and \emph{closed} when it ends, so the ground learns of a convergence about two seconds after it starts.

Sending conclusions instead of evidence is not new: edge analytics systems filter frames by relevance~\cite{canel2019scaling,wang2018bandwidth}, cooperative perception sends object lists~\cite{thandavarayan2020generation}, semantic communication transmits task features~\cite{guo2025videoqasc,shao2024task}, and caption or event-log representations let a language model answer video questions~\cite{zhang2024simple,dixit2026semantic}. What is missing is a measurement: \emph{how much} does a symbolic event stream lose or gain against the video it replaces, against simpler things a drone could send, and on emulated links from 1\,Mbps to 1\,kbps? We contribute that measurement as much as a method:

\begin{itemize}[leftmargin=*,itemsep=1pt,topsep=2pt]
\item \textbf{A transmission code for aerial behavior.} Track-level primitives sent as open and closed records of 14--53 bytes, compact mission cards, and a byte-counted packet format with a strict-priority scheduler and 256-byte fragmentation (\cref{sec:method}). Seven primitives are defined and six run onboard; our data tests one of them, convergence, at scale.
\item \textbf{A preregistered protocol with matched readers and strong baselines.} Questions are answered by a text-only language model reading the stream and by the vision fine-tune of the same base model reading H.265 video. Because that video reader turns out to use little of the bitrate, we add baselines that isolate the code from the reader: the video reader on source frames, video on which the ground runs our own perception, and a stream of quantized tracks. Hypotheses were preregistered, and a held-out set with its mission cards stayed sealed until the pipeline was frozen (\cref{sec:protocol}).
\item \textbf{Findings, positive and negative.} On development data the event stream at 0.13\,kbps answers more questions than the video reader does at any rate, though about three quarters of its margin over H.265 at 100\,kbps comes from questions where nothing happened. When the ground runs our perception on the video, video needs about 100\,kbps to support the same answers, and a track stream needs 96 times the stream's bytes (17 times at a loss of 0.03). On the held-out set, tested in two rounds, the stream built from annotated tracks beats H.265 at 100\,kbps, but the stream from onboard perception does not, and it is no better than a reader that receives nothing. Through links from 1\,Mbps down to 0.3\,kbps the event records lose 0.015 of accuracy, an every-frame track stream holds to 1\,kbps, and the event records tolerate 20\% packet loss. On a Jetson Orin Nano the unchanged pipeline runs at 4.5\,fps; at the frame rate it can keep up with, event recall falls from 0.69 to 0.28 (\cref{sec:results}).
\end{itemize}

\section{Related Work}
\label{sec:related}

\paragraph{Semantic and task-oriented communication for video.}
VideoQA-SC~\cite{guo2025videoqasc} is the closest evaluation: it reports video question-answering accuracy against bandwidth for a learned joint source--channel code, with H.264/H.265 baselines. It differs in the code (learned features, not readable records), the receiver (a small trained model, not a language model) and the unit (channel symbols, not bytes). Question-conditioned scene-graph transmission~\cite{liu2025goal} and receiver-requested text extraction~\cite{liu2024receiver} decide \emph{what} to send per question and off-board; our mission card is a standing specification executed onboard. Feature-level task codes~\cite{shao2024task}, embedding streams read by a multimodal language model~\cite{onsu2026semantic}, prompt-conditioned UAV semantic encoders~\cite{joshi2026context} and generative reconstruction from keyframes or text~\cite{li2025goal,wu2026mirage,ma2026lgvsc} keep a pixel- or feature-derived code; ours is symbolic.

\paragraph{Coding for machines and task-aware codecs.}
Video and image coding for machines~\cite{duan2020video,eimon2026emerging,le2021image} trade pixels for task accuracy, and recent work evaluates codecs through a multimodal language model's answers~\cite{tian2024coding,kao2025bridging,zhang2026benchmarking,baid2026blue,xu2026semabr}, as we do. Region-aware encoders such as AccMPEG~\cite{du2022accmpeg} and server-driven streaming~\cite{du2020server} spend bits where a task needs them; our region-preserving baseline is a simple member of this family.

\paragraph{Edge video analytics.}
On-camera filtering~\cite{li2020reducto,canel2019scaling,kang2017noscope}, configuration adaptation~\cite{jiang2018chameleon} and drone offloading~\cite{wang2018bandwidth,chen2015glimpse} reduce what leaves the device but still send frames for analysis elsewhere; our ground-side perception baseline is this design. Cooperative-perception messages~\cite{thandavarayan2020generation,hu2022where2comm,liu2020when2com} and metadata standards for moving-target tracks and analytics events~\cite{misb0903,onvifprofilem} send objects, not pixels, as our track-stream baseline does, and age-of-information work~\cite{maatouk2020age,li2024towards,sun2023optimizing,han2025age} measures freshness, as our timeliness results do.

\paragraph{Language and structure as the video interface.}
Caption-then-LLM pipelines~\cite{zhang2024simple,zeng2023socratic,wang2025videotree,kahatapitiya2025language}, agents with object-centric memories~\cite{fan2024videoagent,wang2024videoagent,wang2023chatvideo} and event graphs built from detection and tracking~\cite{dixit2026semantic} show that language models answer video questions from compact text. Dixit and Liang~\cite{dixit2026semantic} is closest in representation (tracked objects to event logs to an LLM) but counts tokens, not link bytes.

\paragraph{Aerial video.}
VisDrone~\cite{zhu2022detection} and Okutama-Action~\cite{barekatain2017okutama} provide aerial tracks; other aerial datasets target actions, crowds, captions and anomalies~\cite{li2021uavhuman,wen2021detection,mou2020era,bashmal2023capera,jin2022anomaly}, and UAV question-answering benchmarks are image-based or two-frame~\cite{sun2026uavreason,zhang2026uavqa} or, in MM-UAVBench~\cite{dai2025mmuavbench}, include manually annotated event understanding, tracing and prediction. We found no public aerial-video benchmark that asks where, when and how many tracked objects took part in an event, so we generate questions from annotated tracks (\cref{sec:protocol}).

\section{The Behavior-Primitive Stream}
\label{sec:method}

\begin{figure*}[t]
  \centering
  \includegraphics[width=\linewidth]{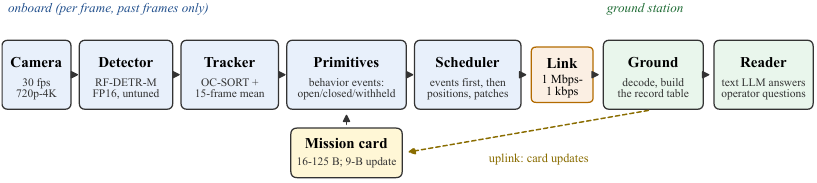}
  \caption{\sys{} onboard and on the ground. Everything left of the link runs per frame. The scheduler sends event records first and fills the remaining capacity with positions, appearance patches and keyframes. The ground turns the received records into a record table, and a language model answers the operator's questions from that table.}
  \label{fig:pipeline}
\end{figure*}

\paragraph{Mission cards.}
A card is the operator's standing request (\cref{fig:pipeline}): free text plus clauses. Each clause names a primitive, the subject classes, a region (a set of cells of a $16{\times}9$ grid on a canonical 1280-pixel-wide frame, or anywhere), a window length (45 to 90 frames in our development cards), a magnitude threshold (absolute, or relative to the scene's own baseline) and an uncertainty threshold. Both ends hold the card. Its human-readable JSON is 7.5--9.8\,kB; the binary form sent over the link is 16--125\,bytes, and changing one clause parameter in flight costs a 9-byte update that takes effect at the next window. Our byte counts charge the JSON form.

\paragraph{Onboard perception.}
A real-time detection transformer (RF-DETR-M~\cite{robinson2026rfdetr}, COCO weights, no fine-tuning, $736{\times}1280$ input, half precision, score $\ge0.3$) feeds OC-SORT~\cite{cao2023ocsort} with one tracker per class group. Boxes are smoothed with a trailing 15-frame mean. Tracker settings were selected on development data over a declared 12-setting grid (supplement); detector weights were never tuned.

\paragraph{Primitives.}
\cref{tab:primitives} lists the seven primitives. Each is a statistic over the tracks inside the clause's region in a sliding window, compared with the clause's threshold, either absolutely or as a robust $z$ score against the scene's other windows (median and median absolute deviation). Windows that exceed the threshold are merged into events. Positions are grid cells; time is frames. Under a moving camera only \texttt{approach} (convergence) is asked, because it depends on distances between tracks and not on the camera's motion. Six primitives run onboard. \texttt{persistent\_change} has no onboard detector: its test objects are synthetic insertions, and a stream from onboard perception carries no record for it.

\begin{table}[t]
  \centering\small
  \setlength{\tabcolsep}{3pt}
  \begin{tabular}{@{}lp{0.72\linewidth}@{}}
    \toprule
    primitive & statistic in the window (canonical pixels, 30\,fps) \\
    \midrule
    dwell & tracks with speed $<0.3$\,px/frame and net displacement $\le8$\,px, among $\ge3$ moving neighbors \\
    density change & subject-class tracks present in the region \\
    birth after quiet & new tracks after $\ge60$ frames at or below the 25th-percentile occupancy \\
    flow deviation & tracks moving $\ge0.5$\,px/frame at $>90^\circ$ to the baseline flow direction, when that flow is coherent (mean resultant length $\ge0.5$) \\
    approach & largest group of $\ge3$ tracks whose mutual distances close by $\ge30\%$ and $\ge20$\,px, ending within 120\,px \\
    persistent change & an object that appears and stays (synthetic insertions; no onboard detector) \\
    access-rate change & entries and exits across the region boundary \\
    \bottomrule
  \end{tabular}
  \caption{The seven primitives. Thresholds are the preregistered values, except the 20-px closing distance of \texttt{approach}, which was added after the first development result and is therefore reported both ways (supplement).}
  \label{tab:primitives}
\end{table}

\paragraph{Records.}
An event produces an \emph{open} record at the first window that exceeds the threshold and a \emph{closed} record when it ends. A record is 13 bytes plus one byte per grid cell it lists (14--53 bytes on development data, 25 on average): clause, primitive, frame span, cells, magnitude, a confidence and an uncertainty level (uncalibrated, and not used by the reader), and the decision rule used. When a clause cannot decide (for example, too few earlier windows), the drone sends a \emph{withheld} record with a reason code, so the ground can tell ``nothing happened'' from ``could not tell''.

\paragraph{Offline and causal decisions.}
The engine has two modes. In \emph{causal} mode a window is compared only with windows that ended before it began; this is what a drone can send live. The per-frame onboard implementation gives exactly the same 98 development events as a whole-sequence run. In \emph{offline} mode a window is compared with all other windows of the sequence, as the question generator's labeller does (\cref{sec:protocol}). The offline mode is the preregistered primary row because it differs from the labeller only in its input tracks; we report the causal row beside it everywhere.

\paragraph{The full stream and its scheduler.}
Besides event records the drone can send quantized track positions (a cell plus a 4-bit offset and size), a camera homography, an appearance patch (JPEG) the first time a track is seen, and a background keyframe (JPEG) when an event is first noticed or the camera view is lost. Every packet has a 12-byte header and is counted byte for byte. A strict-priority scheduler sends event records before positions, and positions before patches and keyframes: semantics first, pixels later. Packets larger than 256 bytes are cut into fragments, so that an event record waits for at most one fragment. On development data this \emph{full stream} averages 64\,kbps, of which event records are about 0.02\%. Below that rate the stream degrades gracefully, in that order: patches and keyframes go first, then positions, and event records are never dropped. All accuracy results in this paper use the \emph{event stream} alone: the event records plus the mission cards, 0.13\,kbps on development data.

\paragraph{The ground.}
The ground decodes the bytes and renders the event records as a \emph{record table}: cell ids become position names, frame spans become time bins, and each clause gets a count, a first time and an order. The table costs no link bytes. We introduced it after a first development run showed that a 7B reader extracts little from raw records, and we report its effect (\cref{sec:results}).

\section{Evaluation Protocol}
\label{sec:protocol}

\paragraph{Data and split.}
We use VisDrone2019-VID~\cite{zhu2022detection} (80 sequences, hover and moving cameras) and Okutama-Action~\cite{barekatain2017okutama} (43 sequences, a moving camera over one sports ground). VisDrone was split 48/32 at sequence level before any event was derived, with sequences of one flight kept together. The 32 sequences are the \emph{held-out set}. Okutama is a single location and goes wholly to development, so development results come mostly from one scene.

\paragraph{Cards and rounds.}
A card is \emph{bound} to the sequences it applies to. Development has four cards (unexpected stop; crowd dynamics; appearance and residue; convergence under a moving camera), drafted with a language model and bound to 47 sequences (43 Okutama, 4 VisDrone). The held-out cards were drafted by other language models from a written brief, reviewed by one author and hashed; the pipeline was developed without access to them. \emph{Round~1} used three cards sealed before any perception code existed; one is derived from a development card and is excluded. They produced questions on only three sequences. We therefore registered \emph{round~2} before its cards existed: five new cards for the 29 held-out sequences that no card bound yet. All 29 have a moving camera, for which the card rules allow only \texttt{approach}; 27 of them produced questions.

\paragraph{Questions.}
The benchmark's events are defined by the same written rules that the stream implements. A question generator that reads only annotations applies the card's rules to the annotated tracks and asks multiple-choice questions about the resulting events, in six families. Presence and nothing-happened questions have two options; localization, counting, timing and ordering have four. A nothing-happened question asks about a frame span in which the event did not occur, so its answer is always no: ``Between frame 454 and frame 513, was it the case that several tracks converged on one spot in the frame?''. A \emph{perfect-perception} stream carries the generator's events, an oracle for these questions; a \emph{real-perception} stream carries the events of the onboard pipeline. The onboard engine re-implements the rules and, given annotated tracks, reproduces all 125 development events of the six onboard primitives exactly, so the two streams differ in perception, not in definition. Each development question is asked in three phrasings (1,638 questions); the held-out set uses two further phrasings per family, written with the first three and never used in development. Of the development questions, \textbf{84.1\% concern \texttt{approach}} because Okutama's staged convergences dominate, 47.1\% are nothing-happened, the 261 questions on other primitives come from the four VisDrone sequences, and \texttt{flow\_deviation} produced none.

\paragraph{Readers.}
The text reader is Olmo-3-7B-Instruct~\cite{olmo2025olmo3}. The video reader is Molmo2-O-7B~\cite{clark2026molmo2}, the vision fine-tune of the same base model, so that the two share a language model. The text reader sees the mission card and the record table of the whole sequence. The video reader sees the card and 16 frames ($378{\times}378$ each), sampled evenly from the frames that arrived within the question's frame span. The \emph{no-stream reader} is the text reader given the question only; it measures what priors and wording give away.

\paragraph{Baselines.}
\emph{Plain H.265} is a \emph{ladder} of six \emph{rungs}, each a target rate with a scale and a frame rate, because below about 159\,kbps libx265~\cite{sullivan2012hevc,x265} stops following the requested bitrate at full resolution. A rung is named by its target in kbps: t1000 (full scale, 30\,fps), t300 (full, 30), t100 (half, 30), t30 (half, 5), t10 (quarter, 2) and t1 (quarter, 1). We report achieved rates. \emph{Low-latency H.265} uses zero-latency tuning, intra refresh and a rate cap at the link rate. \emph{Region-preserving H.265} blurs the background before encoding and keeps every onboard track box at full resolution, in the spirit of task-aware codecs~\cite{du2022accmpeg,baid2026blue}. \emph{Onboard captions}: Molmo2-O-7B captions every 2-second segment, generically or conditioned on the mission cards, and the text reader answers from the caption log~\cite{zhang2024simple}. These three were added after the freeze and were not tuned.

Three further post-freeze baselines separate the code from the reader (declared before they were computed; run on development data and, as exploratory analyses after its registered results, on round~2; supplement). \emph{Source frames}: the video reader on the datasets' own frames, with no video codec in between; this is its ceiling. \emph{Ground-side perception}: each rung is decoded on the ground, the same detector, tracker, smoother and engine run on the decoded frames, and the text reader answers from the same record table. This asks what the bits of video buy when the ground has our perception. Because the ladder's lower rungs also cut the frame rate, four added rungs keep 30\,fps: h60 and h30 (half scale, targets of 60 and 30\,kbps) and q30 and q10 (quarter scale). \emph{Track stream}: the drone sends each track's quantized position (5-px steps) for the classes its cards name, every frame or every sixth frame, and the ground computes the events. This asks whether events need to be decided onboard. An \emph{empty-table} control gives the text reader the cards and a record table without records.

\paragraph{Metrics and hypotheses.}
Accuracy is reported with 95\% intervals from a bootstrap that resamples sequences (2,000 resamples)~\cite{field2007bootstrapping}; development intervals are optimistic, since 43 of the 47 sequences share one location. The preregistered hypotheses are: H1, the event stream answers at least as well as t100 (it \emph{holds} when the interval of the difference lies above zero); H2, at equal accuracy the ladder needs at least 10$\times$ the stream's bytes; H3, t1000 is more accurate than t1. The \emph{byte reduction at equal accuracy} divides the ladder's bytes at the stream's accuracy (interpolated over achieved rates) by the stream's bytes; when the stream is more accurate than every rung, only a lower bound (top-rung bytes over stream bytes) is reported. Event-stream bytes count one record per event with its packet header, plus the mission cards once in JSON; with the open and withheld records also sent, the rate is 0.136 instead of 0.126\,kbps. Video rows are not charged for the card. Timeliness is the delay from onboard emission to the last byte on the ground and, for video, the age of the newest fully received frame~\cite{kaul2012realtime}.

\paragraph{Link emulation.}
A link of rate $R$ carries one packet or fragment at a time for $\text{bytes}\times8/R$ seconds, optionally with independent loss, and no retransmission; propagation delay is ignored. Video is sent first in first out, each frame entering at its capture time. By default libx265 writes a 2.3\,kB information message at every key frame, which a live system would disable. Timing and low-latency rates exclude it; the registered ladder bytes keep it, which overstates the H2 ratios by at most a few percent (supplement).

\paragraph{Preregistration.}
Hypotheses, thresholds, the primary comparison (event stream against t100) and the byte accounting were fixed and hashed before any perception code existed; later changes are numbered amendments. H1 and H2 were registered for the perfect-perception stream, and an amendment made before the held-out set was opened registered the same test for the real-perception stream. The pipeline was frozen (repository tag and file hashes) before the held-out set was opened, every access was logged, and each round was run once. Tracker settings, the smoother, the record table and the video reader's 16 frames were chosen on development data before the freeze. Three small changes to frozen files (supplement) change no held-out number. \emph{Post-freeze} rows were declared after the freeze and run on development data (the caption and region-preserving rows of round~2 were registered with that round); \emph{exploratory} analyses were not registered.

\section{Results}
\label{sec:results}

\begin{table*}[t]
  \centering\small
  \begin{tabular}{@{}lrcccc@{}}
\toprule
row & kbps & accuracy {\scriptsize[95\% CI]} & event & nothing & bold row $-$ row \\
\midrule
no-stream reader (question only) & -- & 0.574 {\scriptsize[0.546, 0.602]} & 0.349 & 0.826 & +0.169 {\scriptsize[+0.134, +0.204]} \\
\midrule
\multicolumn{6}{@{}l}{\emph{video reader on}} \\
\quad source frames (no codec) & source & 0.647 {\scriptsize[0.606, 0.687]} & 0.597 & 0.702 & +0.096 {\scriptsize[+0.049, +0.145]} \\
\quad H.265 t1000 & 998.0 & 0.645 {\scriptsize[0.608, 0.685]} & 0.558 & 0.743 & +0.098 {\scriptsize[+0.050, +0.146]} \\
\quad H.265 t100 & 102.9 & 0.596 {\scriptsize[0.558, 0.635]} & 0.592 & 0.601 & +0.147 {\scriptsize[+0.100, +0.195]} \\
\quad H.265 t10 & 10.8 & 0.583 {\scriptsize[0.548, 0.616]} & 0.596 & 0.568 & +0.160 {\scriptsize[+0.118, +0.207]} \\
\quad H.265 t1 & 2.3 & 0.587 {\scriptsize[0.553, 0.619]} & 0.436 & 0.757 & +0.156 {\scriptsize[+0.117, +0.201]} \\
\midrule
\multicolumn{6}{@{}l}{\emph{text reader on records the ground computes from}} \\
\quad H.265 t1000, decoded & 998.0 & 0.726 {\scriptsize[0.700, 0.755]} & 0.644 & 0.820 & +0.016 {\scriptsize[-0.010, +0.043]} \\
\quad H.265 t100, decoded & 102.9 & 0.727 {\scriptsize[0.690, 0.767]} & 0.625 & 0.842 & +0.016 {\scriptsize[-0.026, +0.052]} \\
\quad H.265 h60, decoded & 63.6 & 0.692 {\scriptsize[0.657, 0.731]} & 0.539 & 0.864 & +0.051 {\scriptsize[+0.016, +0.084]} \\
\quad H.265 q30, decoded & 33.3 & 0.680 {\scriptsize[0.657, 0.706]} & 0.469 & 0.917 & +0.063 {\scriptsize[+0.036, +0.088]} \\
\quad H.265 t30, decoded & 31.3 & 0.662 {\scriptsize[0.619, 0.701]} & 0.421 & 0.933 & +0.081 {\scriptsize[+0.039, +0.129]} \\
\quad H.265 q10, decoded & 13.1 & 0.656 {\scriptsize[0.633, 0.682]} & 0.381 & 0.966 & +0.087 {\scriptsize[+0.058, +0.116]} \\
\quad track stream, every frame & 12.2 & 0.731 {\scriptsize[0.704, 0.759]} & 0.660 & 0.812 & +0.012 {\scriptsize[-0.008, +0.034]} \\
\quad track stream, every 6th frame & 2.1 & 0.717 {\scriptsize[0.688, 0.748]} & 0.604 & 0.844 & +0.026 {\scriptsize[+0.001, +0.051]} \\
\midrule
\multicolumn{6}{@{}l}{\emph{text reader on what the drone sends}} \\
\quad captions, generic & 1.2 & 0.485 {\scriptsize[0.460, 0.509]} & 0.416 & 0.562 & +0.258 {\scriptsize[+0.222, +0.295]} \\
\quad captions, mission-conditioned & 0.576 & 0.454 {\scriptsize[0.422, 0.486]} & 0.472 & 0.435 & +0.289 {\scriptsize[+0.243, +0.332]} \\
\quad \sys{}, real perception, raw records & 0.126 & 0.617 {\scriptsize[0.587, 0.647]} & 0.414 & 0.846 & +0.126 {\scriptsize[+0.097, +0.157]} \\
\quad \sys{}, real perception, causal & 0.125 & 0.739 {\scriptsize[0.712, 0.767]} & 0.649 & 0.839 & +0.004 {\scriptsize[-0.009, +0.015]} \\
\quad \textbf{\sys{}, real perception} & 0.126 & 0.743 {\scriptsize[0.717, 0.770]} & 0.667 & 0.829 & -- \\
\quad \sys{}, perfect perception & 0.129 & 0.800 {\scriptsize[0.772, 0.830]} & 0.809 & 0.790 & -0.057 {\scriptsize[-0.086, -0.032]} \\
\bottomrule
\end{tabular}

  \caption{Development set (1,638 questions, 47 sequences; the onboard pipeline was tuned on this set). \sys{} rows are the event stream; the bold row is the primary one (real perception, offline mode). ``kbps'' counts every byte sent, pooled over all sequences; the \sys{}, track-stream and mission-conditioned caption rows include the four mission cards in JSON. ``event'' and ``nothing'': accuracy on the 867 questions of the five families that ask about an event, and on the 771 nothing-happened questions. Last column: accuracy of the bold row minus this row's (positive: the bold row is better). ``Decoded'' rows are the ground-side perception baseline (h60, q30 and q10 keep 30\,fps). Only the no-stream, plain H.265 and \sys{} rows were in the design before the freeze; all other rows were added after it and were not tuned.}
  \label{tab:dev}
\end{table*}

\begin{figure}[t]
  \centering
  \includegraphics[width=\linewidth]{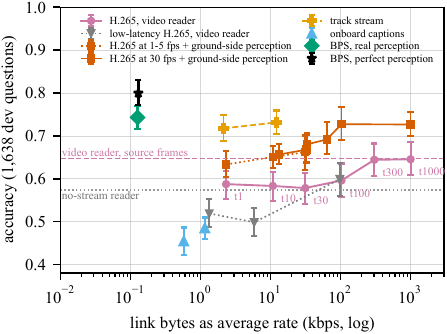}
  \caption{Accuracy against link rate on the development set (95\% intervals, sequences resampled). With the video reader (all six rungs), accuracy barely depends on the rate. With ground-side perception, video matches the event stream at about 100\,kbps, some 800 times the stream's rate, and falls behind at 64\,kbps and below, at 30\,fps as at lower frame rates. Captions lie below the no-stream reader.}
  \label{fig:rate}
\end{figure}

\subsection{Development set}
\cref{tab:dev} and \cref{fig:rate} give the development results. With real perception the event stream answers 0.743 of the questions at 0.13\,kbps. The no-stream reader answers 0.574. The video reader answers 0.596 at t100 (the stream is $+0.147$ [$+0.100$, $+0.195$] higher) and 0.645 at t1000. On the source frames it answers 0.647: the ladder is nearly flat because the reader, at 16 frames of $378{\times}378$, cannot use more bits. The stream is more accurate than every rung, so the byte comparison of H2 is only a bound, at least 7,900$\times$ against t1000. H3 holds (t1000 over t1: $+0.058$ [$+0.024$, $+0.094$]).

\paragraph{Where the margin comes from.}
Nothing-happened questions are 47\% of the bank, and on them the stream (0.829) only equals the no-stream reader (0.826), while the video reader answers ``yes'' too often (0.601 at t100). On the 867 event questions (the other five families), the stream scores 0.667, against 0.349 for no stream ($+0.317$ [$+0.263$, $+0.370$]), 0.592 for t100 ($+0.075$ [$+0.004$, $+0.147$]) and 0.597 for source frames ($+0.069$ [$-0.011$, $+0.151$]). About three quarters of the pooled margin over t100 thus comes from nothing-happened questions, and on event questions the stream's lead over the video reader is small. Perfect perception reaches 0.809 on event questions (by family: supplement). The causal stream is not distinguishable from the offline one ($+0.004$ [$-0.009$, $+0.015$]).

\paragraph{What video buys a ground with the same perception.}
The ground-side perception baseline replaces the video reader with our own perception. From t1000 it finds as many of the annotated events as the drone does (recall 0.68, precision 0.75, against 0.67 and 0.79 onboard, all in offline mode), and the text reader answers 0.726. From t100 it recovers slightly fewer (recall 0.58) and answers 0.727. Neither differs measurably from the event stream ($+0.016$ [$-0.010$, $+0.043$] and $+0.016$ [$-0.026$, $+0.052$], respectively). Below t100 the ladder also cuts the frame rate, so we added four rungs that keep 30\,fps. At 64\,kbps (h60) recall is 0.42 and accuracy 0.692, and the stream is $+0.051$ [$+0.016$, $+0.084$] higher. At 11--33\,kbps, at either frame rate, recall is at most 0.26 and accuracy 0.651--0.680. With the same perception on both sides, the event stream at 0.13\,kbps is therefore not distinguishable from video at about 100\,kbps, a rate ratio of about 800, and video at 64\,kbps or less supports them less well.

\paragraph{Does the drone need to decide events?}
The track stream sends the tracks and lets the ground compute the events. Sent every frame (12.2\,kbps) it gives the same answers as the event stream within noise (0.731; difference $+0.012$ [$-0.008$, $+0.034$]) and similar events (recall 0.69, precision 0.74). Sent every sixth frame (2.1\,kbps) it loses events (recall 0.60) and 0.026 [0.001, 0.051] of accuracy. Deciding events onboard is thus not needed for accuracy; it buys bytes. With the mission cards counted on both sides, the event stream is 96 times smaller than the every-frame track stream and 17 times smaller than the every-sixth-frame one, or 9 and 4 times smaller under a whole-sequence compression bound in the track stream's favor (supplement). Without the cards, which the ground wrote, the event records are 4.3\,kB against 3,989 and 667\,kB of track packets.

\paragraph{The record table and the other baselines.}
Given the raw records instead of the record table, the text reader drops to 0.617, and to 0.414 on event questions: a 7B model does not reliably turn cell ids and frame spans into positions and time bins. Given the cards and an empty record table it scores 0.623, so the cards and the table's framing explain 0.049 [0.020, 0.078] of the margin over the no-stream reader and the records the rest. Region-preserving H.265 scored below plain t100 with the video reader ($-0.022$ [$-0.045$, $-0.002$]). Onboard captions scored below the no-stream reader, by 0.089 (generic) and 0.120 (mission-conditioned): the captioner reports gatherings that did not happen.

\subsection{Held-out confirmatory test}
\begin{table}[t]
  \centering\small
  \setlength{\tabcolsep}{3pt}
  \resizebox{\linewidth}{!}{\begin{tabular}{@{}lcccc@{}}
\toprule
row & accuracy {\scriptsize[95\% CI]} & event$^\dagger$ & nothing$^\dagger$ & row $-$ t100 \\
\midrule
no-stream reader & 0.626 {\scriptsize[0.574, 0.679]} & 0.367 & 0.923 & -- \\
\midrule
H.265 t1000 & 0.647 {\scriptsize[0.585, 0.707]} & 0.610 & 0.688 & -- \\
H.265 t100 & 0.603 {\scriptsize[0.540, 0.668]} & 0.619 & 0.585 & -- \\
H.265 t10 & 0.531 {\scriptsize[0.473, 0.598]} & 0.572 & 0.485 & -- \\
\midrule
region-preserving t100 & 0.566 {\scriptsize[0.494, 0.646]} & 0.604 & 0.523 & -0.037 {\scriptsize[-0.068, -0.003]} \\
captions, generic & 0.435 {\scriptsize[0.383, 0.492]} & 0.435 & 0.436 & -0.168 {\scriptsize[-0.223, -0.111]} \\
captions, mission-cond. & 0.472 {\scriptsize[0.412, 0.544]} & 0.462 & 0.485 & -0.131 {\scriptsize[-0.189, -0.072]} \\
\midrule
\sys{}, perfect perception & 0.737 {\scriptsize[0.693, 0.781]} & 0.840 & 0.619 & +0.133 {\scriptsize[+0.069, +0.203]} \\
\sys{}, real perception & 0.614 {\scriptsize[0.564, 0.667]} & 0.572 & 0.662 & +0.011 {\scriptsize[-0.038, +0.063]} \\
\sys{}, real, causal & 0.588 {\scriptsize[0.539, 0.637]} & 0.505 & 0.683 & -0.016 {\scriptsize[-0.080, +0.049]} \\
\midrule
\sys{}, real, rule-based reader$^\dagger$ & 0.704 {\scriptsize[0.640, 0.769]} & 0.536 & 0.897 & +0.101 {\scriptsize[+0.032, +0.166]} \\
\sys{}, real, causal, rule-based$^\dagger$ & 0.649 {\scriptsize[0.587, 0.712]} & 0.419 & 0.912 & +0.046 {\scriptsize[-0.034, +0.120]} \\
\bottomrule
\end{tabular}
}
  \caption{Round 2 of the held-out confirmatory test: five new cards, 832 questions on 27 sequences, held-out phrasings, five question families (no ordering), all about \texttt{approach}; run once after the freeze. Last column: the row's accuracy minus t100's (positive: the row beats t100); for the first two \sys{} rows it is the preregistered H1 difference. The causal, region-preserving and caption rows are registered secondary rows. $^\dagger$Exploratory, not registered: the split into the 444 event and 388 nothing-happened questions, and the rule-based reader, which looks answers up in the same records.}
  \label{tab:cf}
\end{table}

\paragraph{Rounds.} Round~1 bound only three sequences with questions and could detect no difference under about $\pm0.25$ (supplement). Round~2 (\cref{tab:cf}) asks about convergence on 27 sequences with 92 annotated events. With perfect perception H1 holds, $+0.133$ [$+0.069$, $+0.203$] over t100. The stream is also above t1000 (0.737 vs.\ 0.647), so H2 holds as a bound (at least 2,532$\times$ fewer bytes). With real perception the preregistered comparison is not significant, $+0.011$ [$-0.038$, $+0.063$], and neither is the causal stream's, $-0.016$ [$-0.080$, $+0.049$]. The real-perception stream is also no better than the no-stream reader ($-0.012$ [$-0.068$, $+0.041$]), so its formal byte reduction of 1,164$\times$ says little. The registered primary test is therefore positive only with annotated tracks. On development phrasings over the same sequences (a registered secondary row) the real-perception stream beats t100 by $+0.079$ [$+0.028$, $+0.134$].

\paragraph{Where the held-out loss is (exploratory).}
Two things differ from development. The first is perception: real minus perfect is $-0.123$ [$-0.174$, $-0.076$], and the pipeline found 0.58 of the 92 annotated events. The second is the reader. On the 388 nothing-happened questions the no-stream reader scores 0.923. Given the cards and a record table, the same model scores 0.662 with real perception, 0.619 with perfect perception, and 0.585 when the table is empty: with the round-2 cards and phrasings, the cards and the table alone cost it its correct ``no'' (development, empty table: 0.971). Against that empty table the real-perception records add $+0.129$ [$+0.087$, $+0.173$] over all questions, so the registered null is the sum of a framing loss and a gain from the records. A rule-based reader that looks the answer up in the same records scores 0.704 [0.640, 0.769] with real perception, $+0.078$ [$+0.010$, $+0.140$] over no stream and $+0.101$ [$+0.032$, $+0.166$] over t100. The added baselines rank as on development, one rung higher: ground-side perception matches the stream from t1000 ($+0.016$ [$-0.026$, $+0.061$]) but not from t100 (the stream is $+0.069$ [$+0.018$, $+0.114$] higher), and the track stream sent every frame matches it at 49\,kbps (supplement). These analyses followed the registered results: they locate the loss but do not establish H1.

\subsection{Timeliness over constrained links}

Every development sequence was sent over each emulated link (table in the supplement). Down to 10\,kbps every event record arrives within half a second of its emission. Low-latency H.265 is also timely: at 10\,kbps its newest frame is under a second old 98\% of the time. But the ground answers 0.739 from the causal records and 0.498 from those frames. At 1\,kbps only 27\% of records arrive within 0.5\,s (median 0.66\,s), while the newest low-latency frame is a median 5.9\,s old. Opening records early matters: the first notice of a convergence reaches the ground 1.97\,s after the event begins, against 6.47\,s when only closed records are sent. Loss costs little: with 20\% of the event packets lost the causal stream scores 0.736, and with 40\%, 0.722 (one realization; rule-based reader, mean of 200: 0.727 and 0.690, from 0.756). These delays exclude onboard compute (\cref{sec:compute}).

\paragraph{What survives a narrow link.}
\cref{tab:link} sends each scheme through the link. The event records average 0.02\,kbps, so even 0.3\,kbps carries them: accuracy goes from 0.739 to 0.724, and what is lost are records emitted in a sequence's last 0.3\,s. A track stream in which a newer packet replaces an unsent one also holds down to 1\,kbps when sent every frame (0.712; the event records are $+0.015$ [$-0.010$, $+0.038$] higher), less well when sent every sixth frame ($+0.032$ [$+0.001$, $+0.062$]). At 0.3\,kbps both fall to 0.638, near the empty table. Video with ground-side perception needs about 100\,kbps. Below 1\,kbps only the event records are left.

\paragraph{Low-latency video inside the stream.}
The stream's patches and keyframes do not fit links below about 100\,kbps, so after the freeze we carried low-latency H.265 \emph{inside} the stream, behind the records and the positions (\cref{fig:teaser}b; development only; supplement). At 10\,kbps boxes at the tracked objects' latest positions then cover 0.67 of the annotated objects (low-latency H.265 alone: 0.02), by construction in the stream's favor; video alone is ahead on some measures (supplement). No reader uses the picture.

\subsection{Onboard compute}
\label{sec:compute}
\begin{table}[t]
  \centering\small
  \setlength{\tabcolsep}{4pt}
  \resizebox{\linewidth}{!}{\begin{tabular}{@{}lccccc@{}}
\toprule
link rate (kbps) & 1000 & 100 & 10 & 1 & 0.3 \\
\midrule
\sys{} event records & 0.739 & 0.739 & 0.741 & 0.726 & 0.724 \\
track stream, every frame & 0.728 & 0.729 & 0.724 & 0.712 & 0.638 \\
track stream, every 6th frame & 0.722 & 0.722 & 0.720 & 0.695 & 0.638 \\
H.265, ground-side perception & 0.726 & 0.726 & 0.651 & -- & -- \\
H.265, video reader & 0.645 & 0.596 & 0.587 & -- & -- \\
\bottomrule
\end{tabular}
}
  \caption{Accuracy on the development questions from what reached the ground before the sequence ended, by link rate (causal mode, no loss; post-freeze). Event records are sent alone. Track packets: a newer one replaces an unsent older one. Video rows: the best rung whose rate is at most 10\% above the link rate; ``--'': no rung fits. For reference, the cards with an empty table score 0.623 and the question alone 0.574.}
  \label{tab:link}
\end{table}

On a laptop GPU (RTX 5090, half precision) the pipeline takes 28\,ms per 720p frame on average (36\,fps). Unchanged on a Jetson Orin Nano 8\,GB developer kit (25\,W mode) it takes 170\,ms (5.9\,fps), and 261--307\,ms on 4K and dense 2.7K sequences (4.5\,fps over all frames). Its boxes and events match the laptop's almost exactly. To keep up at 30\,fps input, the board must skip to every seventh frame (4.3\,fps); the supplement simulates such rates. From 30 to 4.3\,fps event recall falls from 0.69 to 0.28, while accuracy falls only from 0.739 to 0.675 and stays above t100 ($+0.079$ [$+0.028$, $+0.131$]). Pooled accuracy hides the loss: as fewer events are reported, nothing-happened questions improve (0.84 to 0.94) while localization falls from 0.74 to 0.51. On this board the pipeline misses most convergences; it needs a lighter detector.

\section{Limitations and Conclusion}
\label{sec:conclusion}

\paragraph{Limitations.}
The evidence is narrower than the method. (i) 84\% of the development questions concern convergence, and 43 of the 47 development sequences come from one sports ground, where the pipeline was tuned; the other primitives have 15--93 questions on four sequences. Confidence fields, withheld records, card updates and the effect of mission conditioning were not evaluated. (ii) Events and questions are generated from annotated tracks by the rules the stream implements, the questions are templated, no external benchmark was used, and the no-stream reader reaches 57\% on development and 63\% on round~2. (iii) On held-out data the real-perception stream beat neither t100 nor the no-stream reader; the analyses that locate the loss are exploratory, and round~2 tests only convergence under moving cameras. (iv) Both readers are 7B models, and the ground-side baseline uses only our perception, which was not fine-tuned on aerial data. (v) The link emulation ignores propagation delay and retransmission; the embedded measurement is one board on a bench, and the simulation of slower processors reuses the laptop's detections. (vi) Mission cards were drafted by language models, not by operators.

\paragraph{Broader impact.}
The method serves surveillance, a dual-use capability: it transmits less about bystanders but makes automated monitoring cheaper, and the ground acts on reports that can be wrong (event precision is about 0.8).

\paragraph{Conclusion.}
A stream of behavior events answers development questions at 0.13\,kbps that video supports only at about 100\,kbps when the ground runs our perception on it, and at no rate when the video reader reads it; sent causally over a 0.3\,kbps link it still answers 0.724. The evidence stops there: a track stream does as well at 96 times the bytes and still works at 1\,kbps, the registered held-out advantage holds only with annotated tracks, and an embedded board misses most events. Better onboard perception and a reader that keeps its priors come next. Nothing in the code is specific to drones, but all of our evidence is.

\section*{LLM Usage Statement}
Generative AI tools were used extensively in this work. \textbf{Code and experiments:} an LLM coding agent (Claude Code, Anthropic) wrote the streaming engine, packet format, scheduler, link emulator and evaluation harness around off-the-shelf detection and tracking models, and designed, preregistered, ran and analysed the experiments with limited step-by-step direction; the authors set the research questions and made the design, stopping and reporting decisions recorded in the preregistration log. \textbf{Ideation:} several method refinements, such as early event-opening records and 256-byte fragmentation, were proposed by the agent from its failure analyses and adopted after author review. \textbf{Figures and tables:} generated by agent-written scripts from the logged results. \textbf{Text and related work:} the agent drafted and edited the paper and the supplementary material and searched for related work; every cited reference was checked against its source. \textbf{Data:} question banks and ground-truth events were generated by agent-written code from the dataset annotations, and the development mission cards were drafted by the agent; the mission cards of the held-out set were drafted by separate models in fresh chats with no access to any result (round 1: Gemini, Google, before any confirmatory result existed; round 2: ChatGPT, OpenAI) and were reviewed by one author. Separately, language and vision-language models are objects of study in this paper (the ground-side readers and the caption baselines, \cref{sec:protocol}). The authors take full responsibility for all content.

{
    \small
    \bibliographystyle{ieeenat_fullname}
    \bibliography{refs}
}
\clearpage
\appendix
\twocolumn[\centering\Large\bfseries Supplementary Material\vspace{1.5em}]
\section{Preregistration and integrity log}
\label{sec:s-prereg}

\paragraph{Order of events.} (1) Evaluation contract only: data, split, development cards, ground-truth events, question banks, H.265 ladder, readers, metrics and byte accounting; every threshold was proposed with a reason and signed; the preregistration and every file of the held-out set were hashed and the repository tagged before any perception, tracking or stream code existed. (2) One correction to the contract before any pipeline existed: frame files of one dataset were sorted as strings, which scrambled the order of the H.265 frames shown to the video reader for 82\% of development questions; the fix moved the primary difference by 0.008 and changed no conclusion (itself a sign of how little this reader takes from its frames). (3) Perception, tracking, primitives and stream built on development data only. (4) Four pre-freeze improvements declared before implementation (half precision, a 12-setting tracker grid with a fixed selection criterion, open records, fragmentation). (5) Pipeline frozen: repository tag plus a SHA-256 over 18 source and configuration files. (6) Amendment registering the real-perception row of the confirmatory test, committed before unsealing. (7) Round 1 of the confirmatory test run once; every read was logged with its amendment id. (8) Caption, region-preserving and compute-constrained experiments declared, then run on development data. (9) Amendment registering a second confirmatory round for the 29 sealed sequences no card bound, committed before any of its cards existed: primary scope the new cards alone, judged by the round-1 rules; secondary rows pooled with round 1, the causal and raw-packet streams, H3, development-style templates, per card, and captions and region-preserving H.265 with the development settings. (10) Round-2 cards drafted, reviewed, validated by script, hashed and tagged. The set chosen first (before validation; drafted by the round-1 model) failed the validation script twice; the registered fallback, a set drafted in parallel by a third model and unopened until then, passed and is the round-2 set. (11) Round 2 run once (SHA-256 of the manifest of the sealed half at that freeze: \texttt{d28aaad9322d458aea8deb5a1ac19b63} \texttt{3fd2853f40b9d2f1160be1ed6268159d}). Its driver (not a frozen file) needed two fixes, each recorded before any output was read: the perception stage tracked only sequences with questions, while the frozen primitive runner needs every bound sequence; and a column name in its scorer collided with one the frozen scorer already returns.

\paragraph{Changes to frozen files.} \emph{\#1}, before any confirmatory result: the primitive runner did not pass the confirmatory access arguments to the frame loader; fixed, development outputs byte-identical. \emph{\#2}, found while rendering a demonstration video: with fragmentation on, the scheduler also replaced position and keyframe packets that had already started, contrary to its declared rule (only unsent packets are replaced); large position packets starved on slow links. Fixed to the declared rule; the development link figures and the 1\,kbps QA check were re-run (pooled changes $\le$0.5 points; QA $-0.011$ [$-0.035$, $+0.011$]). The confirmatory test does not use the link and is unaffected. \emph{\#3}, before any round-2 card existed: the primitive runner gained an argument that selects the round-2 card directory (default unchanged; development events identical row for row).

\paragraph{Disclosed protocol choices made after seeing development results.} The video reader's frames per window were raised from 4 to 16 after the 4-frame curve was flat across the whole ladder (between 0.51 and 0.57 on every rung, on the 180-question subset used for that check); the record table was introduced after a 7B reader extracted little from raw records; the trailing smoother and the tracker setting were selected on development data. The 20-px minimum closing distance of \texttt{approach} was added after the first development result; without it the development stream scores 0.729 [0.702, 0.754] on 1,635 questions, $+0.125$ [$+0.080$, $+0.170$] over t100 (with it: 0.743 and $+0.147$). All were fixed before the held-out set was opened.

\section{Mission cards}
Development cards: D1 unexpected stop (dwell; 4 sequences), D2 crowd dynamics (density change, approach, flow deviation, access-rate change; 2 sequences), D3 appearance and residue (birth after quiet, persistent change; 4 sequences), D4 convergence under a moving camera (approach; 43 sequences); 47 distinct sequences in all, six of them bound by two cards. Each card is JSON with operator text, clauses and a per-threshold rationale; its binary form is 65, 125, 55--63 and 16 bytes. The confirmatory cards were drafted by a different language model from a written brief (task, primitive vocabulary, card schema, a table of the candidate sequences, and raw frames of each: three with the grid drawn on them for hover sequences, one without grid for moving sequences; no annotations and no development result). The two independent cards were written without seeing any development card; the third was written from the structure of one development card with its free text removed, and is therefore excluded from the confirmatory claim. One author reviewed and adopted the cards; they were never shown to the pipeline's developer. Round 2 used the same brief with the round-2 sequence table, drafted by a third model in a fresh chat: five cards, every pool sequence bound by at least one, one designed to fire rarely. Because the authors knew the round-1 result, the review could only check rule compliance and whether an operator would plausibly ask this; any other change was barred. Neither round's text is reproduced here, so that the cards can be reused for a future held-out test.

\section{Question bank}
\begin{table}[h]
\centering\small
\begin{tabular}{@{}lr|lr@{}}
\toprule
family & items & primitive & items \\
\midrule
nothing happened & 771 & approach & 1,377 \\
localization & 372 & birth after quiet & 93 \\
presence & 180 & dwell & 66 \\
counting & 150 & persistent change & 51 \\
timing & 150 & density change & 36 \\
ordering & 15 & access-rate change & 15 \\
 & & flow deviation & 0 \\
\bottomrule
\end{tabular}
\caption{Development bank composition (1,638 items, three paraphrases each question).}
\end{table}
Questions are generated only from annotations (a test fails the build if the generator imports perception or stream code); negatives are asked only where the clause's subject tracks were present; a clip-level ``no'' exists for every clause that never fired where its subjects were present. \texttt{persistent\_change} events are synthetic insertions composited into the frames (the H.265 ladder encodes the composited frames).

\section{Accuracy by question family}
\begin{table*}[t]
\centering\small\setlength{\tabcolsep}{4pt}
\resizebox{\textwidth}{!}{\begin{tabular}{@{}lcccccccccccc@{}}
\toprule
family ($n$) & no stream & source frames & H.265 t100 & region-pres. t100 & t100 ground & t1000 ground & captions gen. & captions miss. & tracks & BPS causal & BPS real & BPS perfect \\
\midrule
presence (180) & 0.30 & 0.63 & 0.66 & 0.74 & 0.73 & 0.72 & 0.61 & 0.71 & 0.74 & 0.72 & 0.73 & 0.84 \\
localization (372) & 0.44 & 0.67 & 0.65 & 0.67 & 0.68 & 0.72 & 0.42 & 0.46 & 0.74 & 0.74 & 0.75 & 0.85 \\
counting (150) & 0.30 & 0.40 & 0.43 & 0.44 & 0.36 & 0.35 & 0.25 & 0.20 & 0.36 & 0.35 & 0.37 & 0.65 \\
timing (150) & 0.21 & 0.56 & 0.49 & 0.50 & 0.61 & 0.64 & 0.31 & 0.47 & 0.66 & 0.61 & 0.66 & 0.83 \\
ordering (15) & 0.73 & 0.80 & 0.80 & 0.80 & 0.73 & 0.87 & 0.73 & 0.80 & 0.80 & 0.93 & 0.80 & 0.73 \\
nothing happened (771) & 0.83 & 0.70 & 0.60 & 0.53 & 0.84 & 0.82 & 0.56 & 0.43 & 0.81 & 0.84 & 0.83 & 0.79 \\
\bottomrule
\end{tabular}
}
\caption{Development accuracy by question family (1,638 questions). ``Source frames'', ``H.265 t100'' and ``region-pres.\ t100'' are read by the video reader; ``t100 ground'' and ``t1000 ground'' are the ground-side perception baseline; ``tracks'' is the track stream sent every frame. Region-preserving H.265 and captions shift answers toward ``yes'': presence rises, nothing-happened falls.}
\end{table*}
Region-preserving H.265 replaces the background by an 8$\times$ down- and up-sampled copy before encoding and keeps every onboard track box, dilated by 25\%, at full resolution, with the plain rung's settings. At 10 and 1\,kbps it scored 0.571 and 0.578 against 0.583 and 0.587 for the plain rungs (differences within their intervals); at 100\,kbps it raised presence from 0.66 to 0.74 while nothing-happened fell from 0.60 to 0.53.

\section{Confirmatory test: rounds, scopes, cards and families}
\begin{table}[h]
\centering\small\setlength{\tabcolsep}{3pt}
\resizebox{\linewidth}{!}{\begin{tabular}{@{}lccc@{}}
\toprule
row & accuracy {\scriptsize[95\% CI]} & row $-$ t100 & byte reduction \\
\midrule
no-stream reader & 0.439 {\scriptsize[0.353, 0.600]} & -- & -- \\
H.265 t1000 & 0.711 {\scriptsize[0.600, 0.794]} & -- & -- \\
H.265 t100 & 0.702 {\scriptsize[0.600, 0.735]} & -- & -- \\
H.265 t10 & 0.500 {\scriptsize[0.235, 0.614]} & -- & -- \\
\sys{}, perfect perception & 0.649 {\scriptsize[0.500, 0.900]} & -0.053 {\scriptsize[-0.235, +0.300]} & 119$\times$ \\
\sys{}, real perception & 0.658 {\scriptsize[0.529, 0.900]} & -0.044 {\scriptsize[-0.206, +0.300]} & 137$\times$ \\
\bottomrule
\end{tabular}
}
\caption{Round 1: the two independent cards, 114 questions on 3 sequences, held-out phrasings, no \texttt{approach} event (seven events: access-rate changes, births after quiet and one density change). With three sequences the intervals are wide and the bootstrap is close to degenerate.}
\end{table}
\begin{table*}[t]
\centering\small\setlength{\tabcolsep}{4pt}
\resizebox{\textwidth}{!}{\begin{tabular}{@{}lrrcccccccc@{}}
\toprule
scope & seq. & items & perfect & real & causal & t1000 & t100 & no stream & real $-$ t100 & real $-$ perfect \\
\midrule
held-out templates (primary) & 27 & 832 & 0.737 & 0.614 & 0.588 & 0.647 & 0.603 & 0.626 & +0.011 {\scriptsize[$-$0.038, +0.063]} & $-$0.123 {\scriptsize[$-$0.174, $-$0.076]} \\
development-style templates & 27 & 1248 & 0.751 & 0.664 & 0.640 & 0.639 & 0.586 & 0.533 & +0.079 {\scriptsize[+0.028, +0.134]} & $-$0.087 {\scriptsize[$-$0.124, $-$0.052]} \\
held-out, without the 20-px rule & 27 & 826 & 0.718 & 0.619 & 0.594 & 0.625 & 0.574 & 0.611 & +0.045 {\scriptsize[$-$0.005, +0.093]} & $-$0.099 {\scriptsize[$-$0.148, $-$0.053]} \\
held-out, pooled with round 1 & 30 & 946 & 0.726 & 0.619 & 0.605 & 0.654 & 0.615 & 0.604 & +0.004 {\scriptsize[$-$0.042, +0.056]} & $-$0.107 {\scriptsize[$-$0.158, $-$0.062]} \\
development-style, pooled with round 1 & 30 & 1419 & 0.738 & 0.662 & 0.652 & 0.646 & 0.598 & 0.532 & +0.063 {\scriptsize[+0.011, +0.119]} & $-$0.076 {\scriptsize[$-$0.112, $-$0.043]} \\
\bottomrule
\end{tabular}
}
\caption{Round 2 by scope (accuracy; differences with 95\% intervals, sequences resampled). Development phrasings ask about the same events with the three development phrasings. ``Without the 20-px rule'' drops the minimum closing distance of \texttt{approach} (20 canonical px), a threshold set after the first development result was known and therefore reported both ways. Pooled rows add round 1's two independent cards.}
\end{table*}
\begin{table}[h]
\centering\scriptsize\setlength{\tabcolsep}{3pt}
\resizebox{\linewidth}{!}{\begin{tabular}{@{}lrrcccccc@{}}
\toprule
card & seq. & items & no stream & t100 & t1000 & perfect & real & causal \\
\midrule
round 1, card 1 & 1 & 28 & 0.54 & 0.96 & 0.89 & 0.75 & 0.75 & 0.79 \\
round 1, card 2 & 3 & 86 & 0.41 & 0.62 & 0.65 & 0.62 & 0.63 & 0.71 \\
round 1, card 3 (derived, excluded) & 2 & 34 & 0.59 & 0.41 & 0.53 & 0.91 & 0.71 & 0.71 \\
round 2, card 1 & 19 & 322 & 0.59 & 0.60 & 0.70 & 0.68 & 0.54 & 0.53 \\
round 2, card 2 & 5 & 122 & 0.61 & 0.70 & 0.74 & 0.80 & 0.61 & 0.60 \\
round 2, card 3 & 5 & 58 & 0.60 & 0.53 & 0.57 & 0.81 & 0.72 & 0.67 \\
round 2, card 4 & 10 & 186 & 0.60 & 0.33 & 0.33 & 0.67 & 0.52 & 0.50 \\
round 2, card 5 & 11 & 144 & 0.77 & 0.91 & 0.90 & 0.88 & 0.85 & 0.79 \\
\bottomrule
\end{tabular}
}
\caption{Accuracy by card on held-out phrasings. Cards share sequences; questions are per card. Round 1, card 3 was derived from a development card and is excluded from every claim.}
\end{table}
\begin{table*}[t]
\centering\small\setlength{\tabcolsep}{4pt}
\resizebox{\textwidth}{!}{\begin{tabular}{@{}lrccccccc@{}}
\toprule
family & items & no stream & t100 & t1000 & perfect & real & real $-$ no stream & real $-$ t100 \\
\midrule
counting & 80 & 0.175 & 0.375 & 0.362 & 0.775 & 0.275 & +0.100 {\scriptsize[-0.026, +0.223]} & -0.100 {\scriptsize[-0.267, +0.054]} \\
localization & 184 & 0.386 & 0.565 & 0.592 & 0.821 & 0.549 & +0.163 {\scriptsize[+0.058, +0.278]} & -0.016 {\scriptsize[-0.142, +0.144]} \\
nothing happened & 388 & 0.923 & 0.585 & 0.688 & 0.619 & 0.662 & -0.260 {\scriptsize[-0.349, -0.169]} & +0.077 {\scriptsize[-0.037, +0.185]} \\
presence & 100 & 0.680 & 0.840 & 0.760 & 0.850 & 0.690 & +0.010 {\scriptsize[-0.104, +0.130]} & -0.150 {\scriptsize[-0.275, -0.022]} \\
timing & 80 & 0.125 & 0.713 & 0.713 & 0.938 & 0.775 & +0.650 {\scriptsize[+0.500, +0.770]} & +0.062 {\scriptsize[-0.066, +0.200]} \\
\midrule
event questions & 444 & 0.367 & 0.619 & 0.610 & 0.840 & 0.572 & +0.205 {\scriptsize[+0.142, +0.273]} & -0.047 {\scriptsize[-0.115, +0.024]} \\
all & 832 & 0.626 & 0.603 & 0.647 & 0.737 & 0.614 & -0.012 {\scriptsize[-0.068, +0.041]} & +0.011 {\scriptsize[-0.038, +0.063]} \\
\bottomrule
\end{tabular}
}
\caption{Round 2 by question family (held-out phrasings; exploratory, not registered). ``Perfect'' and ``real'' are the \sys{} streams read by the language-model reader. Intervals resample sequences (27; 25 for nothing-happened).}
\end{table*}
\paragraph{Registered secondary rows.} On held-out phrasings the no-stream reader is strong (0.626), and the real-perception stream adds nothing measurable over it ($-0.012$ [$-0.068$, $+0.041$]); on development phrasings, where that prior is weaker (0.533), it adds $+0.131$ [$+0.088$, $+0.181$]. H3 (t1000 above t1) holds on round 2, $+0.125$ [$+0.093$, $+0.158$]. Raw records ($-0.025$ [$-0.082$, $+0.032$]) and the causal stream ($-0.016$ [$-0.080$, $+0.049$]) are not distinguishable from t100. Captions (on round 2: generic 1.33\,kbps, mission-conditioned 0.91\,kbps) and region-preserving H.265 at t100 (121\,kbps on round 2) were run with the development settings. The round-2 event stream totals 18.0\,kB with real and 19.0\,kB with perfect perception, the five cards' JSON included. Captions scored below no stream ($-0.191$ [$-0.243$, $-0.141$] generic, $-0.154$ [$-0.220$, $-0.092$] mission-conditioned) and below the real-perception stream ($-0.179$ and $-0.142$, respectively); region-preserving H.265 scored $-0.037$ [$-0.068$, $-0.003$] against plain t100, and the real-perception stream was not distinguishable from it ($+0.048$ [$-0.016$, $+0.108$]). By card, perfect perception beats t100 on four of the five round-2 cards and t1000 on three.

\paragraph{Exploratory analyses (not registered; made after the registered results were seen; no new reader run).} The split by family (the round-2 family table) and a rule-based reader. The rule-based reader is a short program that answers each question family by look-up in the received records (presence: any record of the clause; nothing-happened and counting: records overlapping the question's frame span; localization: the cells of those records; timing: the time bin of the first record; ordering: the order of first records). With perfect perception it scores 1.0 by construction, because the questions are generated from the same events. On development it scores 0.775 [0.721, 0.822] with real perception and 0.756 with the causal stream, against 0.743 and 0.739 for the language-model reader. On round 2 it scores 0.704 [0.640, 0.769] and 0.649 [0.587, 0.712], respectively, against 0.614 and 0.588; by family, with real perception: counting 0.275, localization 0.457, nothing-happened 0.897, presence 0.80, timing 0.65. The language-model reader's loss on round 2 is on nothing-happened questions (0.662 against 0.897), where it gives up the answer ``no'' that it gives without any stream (0.923). The empty-table control below shows that the cards and the table cause this, not the records.

\section{Byte accounting and packet format}
Every packet has a 12-byte header (magic and version, sequence hash, frame index, card index, flags, payload length). Sections: event descriptor (13\,B plus 1\,B per listed cell), withheld record, quantized track positions (a cell id plus 4-bit offsets and size per track), camera homography, ROI patch (JPEG), background keyframe (JPEG). All accuracy numbers use the event stream (event records and cards) only; the full stream (64\,kbps on development data: appearance patches 49.5\%, keyframes 44.8\%, positions and homographies 4.4\%, headers 1.3\%, event records 0.02\%) is what the ground renders for a human. Mission cards are counted once per mission in their JSON form (7.5--9.8\,kB each), which dominates the event stream's byte count on development data; with their binary form the development event stream would be about 9$\times$ smaller.

\section{Reader prompts}
\paragraph{Text reader with the record table.} A development example of the reader's whole input:

\begin{lstlisting}
MISSION CARD "D2_crowd_dynamics": Watch the kerb and the crossing. Tell me if people bunch up or clear out, if a group converges on one spot, if the foot traffic starts moving against the usual direction, or if the rate of people crossing changes sharply. Give me where and when, and roughly how many.
  clause d2c1: density_change -- subjects: pedestrian, people -- region: cells [32, 33, 34, ...] ("the kerb and stalls")
  ... (three more clauses)
SCENE: 348 frames at 30 fps.
A 16 x 9 grid is laid over the frame. A cell id is row * 16 + col, with row 0 the top row and col 0 the left column.
Time bins used below: bin 1 = frames 0-86 (about 0-3 s in); ... bin 4 = frames 261-347 (about 9-12 s in)
SEMANTIC STREAM RECEIVED ON THE GROUND (rendered by the ground station):
  1. clause d2c1 (density_change): frames 145-344 (starts in bin 2), location: the left side, magnitude 3.06667, cells [...]
  2. clause d2c4 (access_rate_change): frames 170-269 (starts in bin 2), location: the bottom centre, magnitude 6, cells [...]
  3. clause d2c2 (approach): frames 225-289 (starts in bin 3), location: the left side, magnitude 3, cells [...]
PER-CLAUSE SUMMARY (whole clip, frames 0-347):
  clause d2c2: happened 1 separate time(s); first at frame 225 (bin 3, ...)
  order of first occurrence: d2c1 -> d2c4 -> d2c2
QUESTION: several tracks converged on one spot in the kerb and stalls - first time, roughly when?
OPTIONS: 0) frames 87-173 ... 3) frames 174-260 (about 6-9 s in)
Answer with the number of the single best option and nothing else. Do not explain.
\end{lstlisting}
\paragraph{Video reader.} ``You are shown 16 frame(s) sampled evenly from frames $a$--$b$ of an aerial video ($n$ frames at 30 fps). They arrived over a $r$ kbps link, so they may be blurred, downscaled or few.'' followed by the same card, grid note and question block. On source frames the second sentence is omitted. The no-stream reader receives the question block only (from ``QUESTION:'' on).

\paragraph{Onboard captioner (generic).} ``These 8 frames are 2.0 seconds of video from a surveillance drone's camera, in time order. Describe what happens for a ground operator who cannot see the video: which objects are present (people, vehicles, two-wheelers) and roughly how many; where they are, using cell ids of a 16 x 9 grid laid over the frame (cell id = row * 16 + col, row 0 at the top, col 0 at the left) or words such as upper-left; and what they do (stay, move, gather, enter, leave, approach each other). At most 60 words.'' The mission-conditioned variant appends the full text of the sequence's cards and ``Report only what matters for these mission cards, with locations as cell ids or words. If nothing relevant happens in these seconds, answer exactly: nothing relevant. At most 60 words.'' Captions average 281 bytes (generic) and 112 bytes (mission; 251 of 1,345 windows ``nothing relevant''); one caption took 0.99\,s on the host GPU.

\section{Baselines added after the freeze}
\begin{table*}[t]
\centering\small\setlength{\tabcolsep}{4pt}
\begin{tabular}{@{}lcccccc@{}}
\toprule
what crosses the link & kbps & events & recall & precision & accuracy {\scriptsize[95\% CI]} & event questions \\
\midrule
\sys{} event stream (onboard events) & 0.126 & 107 & 0.67 & 0.79 & 0.743 {\scriptsize[0.717, 0.770]} & 0.667 \\
\midrule
H.265 t1000 decoded, ground-side perception & 998.0 & 113 & 0.68 & 0.75 & 0.726 {\scriptsize[0.700, 0.755]} & 0.644 \\
H.265 t100 decoded, ground-side perception & 102.9 & 98 & 0.58 & 0.74 & 0.727 {\scriptsize[0.690, 0.767]} & 0.625 \\
H.265 h60 decoded, ground-side perception & 63.6 & 80 & 0.42 & 0.66 & 0.692 {\scriptsize[0.657, 0.731]} & 0.539 \\
H.265 h30 decoded, ground-side perception & 32.6 & 48 & 0.25 & 0.65 & 0.668 {\scriptsize[0.638, 0.701]} & 0.451 \\
H.265 q30 decoded, ground-side perception & 33.3 & 57 & 0.26 & 0.58 & 0.680 {\scriptsize[0.657, 0.706]} & 0.469 \\
H.265 q10 decoded, ground-side perception & 13.1 & 9 & 0.06 & 0.78 & 0.656 {\scriptsize[0.633, 0.682]} & 0.381 \\
H.265 t30 decoded, ground-side perception & 31.3 & 36 & 0.22 & 0.75 & 0.662 {\scriptsize[0.619, 0.701]} & 0.421 \\
H.265 t10 decoded, ground-side perception & 10.8 & 10 & 0.06 & 0.80 & 0.651 {\scriptsize[0.626, 0.677]} & 0.373 \\
H.265 t1 decoded, ground-side perception & 2.3 & 8 & 0.05 & 0.75 & 0.633 {\scriptsize[0.603, 0.665]} & 0.349 \\
\midrule
track stream, every frame & 12.2 (1.10) & 116 & 0.69 & 0.74 & 0.731 {\scriptsize[0.704, 0.759]} & 0.660 \\
track stream, every 6th frame & 2.1 (0.49) & 94 & 0.60 & 0.80 & 0.717 {\scriptsize[0.688, 0.748]} & 0.604 \\
\bottomrule
\end{tabular}

\caption{Development set: what the same text reader answers when the events are computed onboard (first row) or on the ground. Events, recall and precision are against the 125 annotated events of the six onboard primitives, offline mode. For the track streams the rate is that of the packets, cards included; in brackets, the same under an optimistic whole-sequence compression bound.}
\end{table*}
All were declared, with their settings, before they were computed. They were run on development data and, after the registered results of round~2, on its 27 sequences as exploratory analyses (\cref{tab:r2x}).

\paragraph{Source frames.} The video reader's job is unchanged (16 frames sampled evenly in the question's frame span, same card, grid note and question), but the images are the datasets' own JPEG frames, with no video codec in between. It scores 0.647 [0.606, 0.687], against 0.645 at t1000.

\paragraph{Ground-side perception.} Each plain rung is decoded and the frozen detector runs on every decoded frame (same weights, half precision, same call); boxes are scaled back to source pixels. The frozen tracker runs as in the compute-constrained simulation: a decoded frame sits at the source position it was encoded from, \texttt{max\_age} is $\mathrm{round}(30/\text{step})$ for a rung that keeps every step-th frame, \texttt{min\_hits} is 1, and boxes are interpolated linearly between two outputs exactly one step apart. The smoother, the primitive engine, the record table and the text reader are unchanged. The rate is the rung's; the card is not charged. Two caveats favor the onboard stream: the detector was never tuned for any resolution, and the tracker setting was selected on source-resolution development frames.

\paragraph{Track stream.} The drone sends the smoothed box center of every track of a class its cards need (a clause that names no class takes every class; on the development set this is every track), quantized as the stream's position section quantizes it (a cell of 80 canonical pixels plus 4-bit offsets, 5-px steps), 6 bytes per track in a packet with a 12-byte header and a 3-byte section header, every frame or every sixth frame. The ground interpolates linearly between two samples of a track, reconstructs the center of the 5-px bin, and runs the same engine with the same card. A control with unquantized positions every frame reproduces the onboard events exactly (107 offline, 98 causal). Excluding the cards, the event records total 4.3\,kB on the development set, against 125--667\,kB for the every-sixth-frame track stream (compression bound and packets, respectively) and 326--3,989\,kB every frame. Sent causally, the every-sixth-frame stream scores 0.722.

\paragraph{Rungs that keep 30\,fps.} The ladder's rungs below t100 lower the frame rate to 5, 2 and 1\,fps, which by itself removes events from a tracker. Four added rungs keep 30\,fps: half scale with targets of 60 and 30\,kbps (h60, h30) and quarter scale with targets of 30 and 10\,kbps (q30, q10), encoded with the ladder's settings. Their achieved pooled rates are 63.6, 32.6, 33.3 and 13.1\,kbps. Recall falls from 0.58 at t100 to 0.42 at h60, 0.25 at h30, 0.26 at q30 and 0.06 at q10. The 5-fps rung t30 (31.3\,kbps, recall 0.22, accuracy 0.662) is close to the two 30-fps rungs of the same rate (0.668 and 0.680), so between 13 and 33\,kbps the rate, not the frame rate, sets the result.

\paragraph{Controls (development; declared before computing).} \emph{Empty table.} The text reader given the cards and a record table without records scores 0.623 [0.586, 0.659], which is 0.049 [0.020, 0.078] above the no-stream reader (0.313 on event questions and 0.971 on nothing-happened questions; the causal stream: 0.649 and 0.839). \emph{Loss.} Each event packet (the records one card emits at one frame; 96 packets carry the 98 causal records) is lost independently with probability $p$ and not retransmitted. With one realization read by the text reader, accuracy is 0.742, 0.742, 0.736 and 0.722 at $p = 5$, 10, 20 and 40\% (0.739 without loss; at 40\%: $-0.016$ [$-0.041$, $+0.007$]). The rule-based reader, averaged over 200 realizations, scores 0.750, 0.743, 0.727 and 0.690 (0.756 without loss). \emph{Track stream over links} (no loss). Sent every sixth frame the track stream offers 2.0\,kbps (packets only; 2.1 with the cards). At 10\,kbps 99\% of its packets arrive within 0.5\,s. At 1\,kbps, first in first out, 56\% arrive before the sequence ends (median delay 8.9\,s); when a newer packet replaces an unsent one, 54\% arrive and 33\% within 0.5\,s. Sent every frame it offers 12.0\,kbps (12.2 with the cards): at 10\,kbps 29\% of packets arrive within 0.5\,s (77\% with replacement), and at 1\,kbps 10\% arrive. \emph{Question only, video model.} The video reader's model given the question alone scores 0.446 on development and 0.466 on round~2, below the text model given the question alone (0.574 and 0.626).

\begin{table*}[t]
\centering\small\setlength{\tabcolsep}{4pt}
\resizebox{\textwidth}{!}{\begin{tabular}{@{}lcccccccc@{}}
\toprule
what the text reader gets & events & recall & precision & accuracy {\scriptsize[95\% CI]} & event & nothing & onboard stream $-$ row & rule-based \\
\midrule
no stream (question only) & -- & -- & -- & 0.626 {\scriptsize[0.574, 0.679]} & 0.367 & 0.923 & -0.012 {\scriptsize[-0.068, +0.041]} & -- \\
H.265 t100, video reader & -- & -- & -- & 0.603 {\scriptsize[0.540, 0.668]} & 0.619 & 0.585 & +0.011 {\scriptsize[-0.038, +0.063]} & -- \\
\sys{} event stream, onboard events & 70 & 0.58 & 0.76 & 0.614 {\scriptsize[0.564, 0.667]} & 0.572 & 0.662 & -- & 0.704 \\
\midrule
H.265 t1000 decoded, ground-side perception & 71 & 0.55 & 0.72 & 0.599 {\scriptsize[0.551, 0.645]} & 0.570 & 0.631 & +0.016 {\scriptsize[-0.026, +0.061]} & 0.683 \\
H.265 t100 decoded, ground-side perception & 35 & 0.24 & 0.63 & 0.546 {\scriptsize[0.500, 0.596]} & 0.468 & 0.634 & +0.069 {\scriptsize[+0.018, +0.114]} & 0.601 \\
H.265 t30 decoded, ground-side perception & 7 & 0.07 & 0.86 & 0.512 {\scriptsize[0.465, 0.566]} & 0.419 & 0.619 & +0.102 {\scriptsize[+0.053, +0.153]} & 0.531 \\
H.265 t10 decoded, ground-side perception & 0 & -- & -- & 0.486 {\scriptsize[0.440, 0.535]} & 0.396 & 0.588 & +0.129 {\scriptsize[+0.087, +0.174]} & 0.490 \\
H.265 t1 decoded, ground-side perception & 0 & -- & -- & 0.484 {\scriptsize[0.438, 0.535]} & 0.394 & 0.588 & +0.130 {\scriptsize[+0.088, +0.176]} & 0.490 \\
\midrule
track stream, every frame & 70 & 0.55 & 0.73 & 0.601 {\scriptsize[0.547, 0.654]} & 0.572 & 0.634 & +0.013 {\scriptsize[-0.018, +0.051]} & 0.702 \\
track stream, every 6th frame & 58 & 0.48 & 0.76 & 0.577 {\scriptsize[0.532, 0.626]} & 0.545 & 0.613 & +0.037 {\scriptsize[-0.003, +0.082]} & 0.671 \\
\midrule
cards and an empty record table & 0 & -- & -- & 0.486 {\scriptsize[0.440, 0.536]} & 0.399 & 0.585 & +0.129 {\scriptsize[+0.087, +0.173]} & 0.490 \\
\bottomrule
\end{tabular}
}
\caption{Round 2 of the held-out test, exploratory (832 questions, held-out phrasings; run after the registered results, declared before computing). The first three rows are registered runs. Events, recall and precision are against the 92 annotated events, offline mode. ``Rule-based'': the rule-based reader on the same records.}
\label{tab:r2x}
\end{table*}
\paragraph{Round 2, exploratory baselines.} \cref{tab:r2x} repeats the added baselines on round~2. Ground-side perception from t1000 emits about as many events as the drone (71 against 70; recall 0.55 against 0.58) and is not distinguishable from the onboard stream. From t100 it emits 35 and the onboard stream is 0.069 [0.018, 0.114] more accurate; from t30 down it emits almost none. The track stream sent every frame matches the onboard stream at 49.0\,kbps with the cards counted, and sent every sixth frame it is 0.037 [$-0.003$, $+0.082$] lower at 8.5\,kbps (2.6\,kbps under the compression bound); the event stream is 0.36--0.38\,kbps on these sequences. With the cards and an empty table the reader scores 0.486, which is 0.141 [0.098, 0.183] \emph{below} the reader given the question alone: on nothing-happened questions it scores 0.585 against 0.923. The onboard stream is 0.129 [0.087, 0.173] above the empty table. On round~2 the registered comparison with the no-stream reader therefore mixes a loss from the cards and the table with a gain from the records.

\paragraph{What reaches the ground at each link rate (development; declared before computing).} For the link table of the main text each scheme is sent through the link emulator at 1000, 100, 10, 1 and 0.3\,kbps without loss, and the ground answers from what was delivered before the sequence ended, in causal mode. The lowest rate lies below the range the paper set out to study and was added to see where each scheme stops. \emph{Event stream:} the event packets of the frozen stream alone (204 packets; open and withheld records are sent too); the table holds the closed records whose packet arrived: 98, 98, 97, 84 and 83 of 98. The 14 records missing at 1\,kbps were emitted in the last 0.3\,s of a sequence (the 15 at 0.3\,kbps in the last 2.1\,s) and arrive after it ends. In the full stream somewhat fewer event packets arrive before the end (92.6\% against 94.1\% at 1\,kbps). At the same link rate the event stream minus the track stream is $+0.015$ [$-0.010$, $+0.038$] (every frame) and $+0.032$ [$+0.001$, $+0.062$] (every sixth frame) at 1\,kbps, and $+0.086$ [$+0.056$, $+0.116$] at 0.3\,kbps. \emph{Track stream:} the packets of the track-stream baseline, with a newer packet replacing an older unsent one, the scheduler's rule for position packets. Of the packets, 99.5\%, 54\%, 18\% (every sixth frame) and 77\%, 10\%, 3\% (every frame) arrive at 10, 1 and 0.3\,kbps. The ground bridges two consecutive received samples of a track, at most 30 frames apart, only if no packet arrived between them and either they are one sampling step apart or a packet sent between them was missed; with every packet delivered this is the reconstruction of the track-stream baseline, and the event tables are identical. This rule replaced a first one (bridge any two received samples at most 30 frames apart) after the event tables had been built and before any score was computed, because the first rule changed the events even without a link. \emph{Video:} no new run; a cell is the best rung of the ladder or the 30-fps rungs whose pooled rate is at most 10\% above the link rate (t100 at 103\,kbps counts for 100\,kbps, t10 at 10.8\,kbps for 10\,kbps), which favors video; no rung fits 1\,kbps. With the rule-based reader the event stream scores 0.756, 0.731, 0.731 at 10, 1 and 0.3\,kbps, the every-sixth-frame track stream 0.705, 0.679, 0.504 (first in first out: 0.705, 0.656, 0.502) and the every-frame track stream 0.722, 0.700, 0.502 (first in first out: 0.722, 0.496, 0.489). On event questions the language-model reader scores 0.617 and 0.618 with the event stream at 1 and 0.3\,kbps, 0.572 and 0.348 with the every-frame track stream.

\section{Link emulation and scheduler}
\begin{table}[h]
  \centering\small
  \setlength{\tabcolsep}{2.5pt}
  \resizebox{\linewidth}{!}{\begin{tabular}{@{}lccccccc@{}}
\toprule
link & \multicolumn{2}{c}{plain H.265} & \multicolumn{2}{c}{low-latency H.265} & \multicolumn{2}{c}{BPS event records} & captions \\
\cmidrule(lr){2-3}\cmidrule(lr){4-5}\cmidrule(lr){6-7}\cmidrule(l){8-8}
 & age $\le$1\,s & median age & age $\le$1\,s & median age & $\le$0.5\,s & median & median \\
\midrule
1 Mbps & 55\% & 0.7 & 100\% & 0.3 & 100\% & 0.00 & 0.99 \\
100 kbps & 39\% & 1.3 & 100\% & 0.4 & 100\% & 0.00 & 1.00 \\
10 kbps & 1\% & 4.3 & 98\% & 0.5 & 100\% & 0.10 & 1.06 \\
1 kbps & 0\% & 20.0 & 0\% & 5.9 & 27\% & 0.66 & 1.73 \\
\bottomrule
\end{tabular}
}
  \caption{Timeliness on the 47 development sequences over an emulated link without loss. H.265 columns: share of time in which the newest fully received frame is at most 1\,s old, and its median age (s). \sys{} columns: share of event records delivered within 0.5\,s of onboard emission, and the median delay (s). Captions: median delay after the end of their 2-s segment (s), including 0.99\,s of caption generation. The two shares measure different quantities.}
  \label{tab:time}
\end{table}

Rate $R$, one packet or fragment at a time, $\text{bytes}\times8/R$ seconds each, independent loss per packet with no retransmission, no propagation delay. Priority tiers: event and withheld records; positions and homographies; patches and keyframes. A newer position (or keyframe) packet replaces an older one of the same kind that has not started. With fragmentation, a packet over 256 payload bytes is cut into fragments of at most 256 bytes plus a 4-byte header, the scheduler re-decides at every fragment boundary, and a packet is delivered when all its fragments are. Plain H.265 is sent first in first out in file (decode) order, a frame's bytes entering at its capture time; the ground shows the newest frame, in display order, all of whose bytes (and all earlier bytes) have arrived.

\section{Compute-constrained simulation}
\begin{table}[h]
  \centering\small
  \setlength{\tabcolsep}{3pt}
  \resizebox{\linewidth}{!}{\begin{tabular}{@{}lccc@{}}
\toprule
onboard rate & event recall & precision & accuracy {\scriptsize[95\% CI]} \\
\midrule
30 fps (all frames) & 0.69 & 0.82 & 0.739 {\scriptsize[0.712, 0.767]} \\
15 fps (every 2nd frame) & 0.58 & 0.78 & 0.719 {\scriptsize[0.690, 0.750]} \\
10 fps (every 3rd frame) & 0.53 & 0.70 & 0.711 {\scriptsize[0.679, 0.747]} \\
5 fps (every 6th frame) & 0.34 & 0.78 & 0.684 {\scriptsize[0.652, 0.721]} \\
4.3 fps (every 7th frame; Orin Nano) & 0.28 & 0.86 & 0.675 {\scriptsize[0.644, 0.708]} \\
\bottomrule
\end{tabular}
}
  \caption{Slower onboard processors, simulated on the development set: the frozen pipeline sees every $k$-th frame, tracks at that rate and interpolates to 30\,fps. Recall, precision and accuracy are those of the causal real-perception stream (116 annotated events in causal mode). The last row is the rate at which a Jetson Orin Nano keeps up.}
  \label{tab:compute}
\end{table}

The frozen detector output on frames $f \equiv 0 \pmod k$; OC-SORT with \texttt{max\_age} $=\mathrm{round}(30/k)$ and \texttt{min\_hits} 1 on those frames only; boxes mapped back to source indices and linearly interpolated between two outputs of the same track exactly $k$ frames apart; then the unchanged smoother and primitive engine. $k{=}1$ reproduces the frozen tracks exactly. By family at 5\,fps: localization 0.55 (0.74 at 30\,fps), presence 0.56 (0.72), timing 0.41 (0.61), counting 0.20 (0.35), nothing happened 0.92 (0.84). At 4.3\,fps ($k{=}7$): localization 0.51, presence 0.52, timing 0.38, ordering 0.67 (0.93), counting 0.20, nothing happened 0.94.

\paragraph{Embedded measurement (declared before the board runs; development only).} On the laptop GPU the pipeline takes 27.7\,ms per 720p frame on average, 37.7\,ms at 4K and 53.3\,ms on the densest 2.7K sequence. Jetson Orin Nano 8\,GB developer kit, JetPack 6 (L4T R36.4.4, CUDA 12.6), 25\,W mode, desktop off, frames read from the SD card. The frozen code and command line unchanged; RF-DETR-M weights byte-identical to the laptop's; torch 2.8.0 for JetPack 6, which forces NumPy 1.26 and therefore OpenCV 4.11 (laptop: 5.0), used for decoding, colour conversion and the encoder's resize. Three clips (720p, 2.7K, 4K), all frames, the first 20 discarded. Per-stage medians at 720p: detector 125\,ms (laptop 18), tracker 7.3 (4.1), encoder 17 (2.8); at 4K and 2.7K the full-resolution decode, colour conversion and downscale add CPU work, and GPU load averages 36--38\% (54\% at 720p). Power from \texttt{tegrastats} every 100\,ms on the module input rail: 3.39\,W idle (60\,s), 10.64\,W at 720p (7.25\,W above idle; 1.23\,J per frame above idle), 7.4--7.6\,W on the slower 2.7K and 4K clips (4.0--4.2\,W above idle). A run fed at 30\,fps measured a median latency of 91.9\,s against 92.9\,s from a single-server queue model built from the unpaced service times. Compared with the laptop, 98--100\% of tracked boxes match at IoU\,$\ge$\,0.5 (median coordinate difference 0.01--0.07\,px); events are identical at 720p and differ by one on each of the two dense clips. The board's over-current counter rose once, during the last (30-fps-paced) run only. The pooled sustained rate, 4.5\,fps, set $k{=}\lceil 30/4.5\rceil{=}7$ by a rule fixed before the runs, and the $k{=}7$ row was computed after that.

\section{Low-latency video, the encoder's info SEI, and a picture layer}
\paragraph{Low-latency H.265.} After the freeze we added H.265 tuned for live links (zero latency: no B-frames and no look-ahead; intra refresh with a 2-s period; VBV rate cap at the link rate with a 0.5-s buffer), at the ladder's scale and frame rate for each link, plus 1/8-scale variants at 10 and 1\,kbps. It matches the plain ladder's accuracy at 1\,Mbps and 100\,kbps and is worse at 10 and 1\,kbps (0.498 and 0.519 vs.\ 0.583 and 0.587), since without B-frames and look-ahead the same bits buy less quality.

\paragraph{The info SEI (correction).} libx265 writes a user-data-unregistered SEI with its version and full option string (about 2.3\,kB) at every key frame. It is not needed for decoding. We found it while diagnosing a slow first picture: it made up 30\% of the 1\,kbps ladder rung's bytes (5\% at 10\,kbps, 2.4\% at 100\,kbps, 0.25\% at 1\,Mbps) and 63--88\% of the low-latency rungs at 1--10\,kbps, whose 2-s key-frame period repeats it; with it, the low-latency rungs appeared to overshoot their rate by 1.6--11$\times$. We removed it from every stream (decoded frames verified bit-identical) and re-timed all video; the main paper reports the corrected timing. Accuracy is unaffected. The registered ladder bytes keep it; the development iso-utility bound becomes 7,898$\times$ instead of 7,918$\times$ without it.

\paragraph{A picture layer (post-freeze, development only).} The frozen stream sends appearance patches and occasional keyframes, which do not fit links below about 100\,kbps. Every variant below was declared, with its measurements, before it was computed, and all were measured on the 47 development clips against low-latency H.265 alone on the same link (without the info SEI). In order: (1) per-frame JPEG thumbnails refreshed every 0.5\,s (at 10\,kbps only 64-pixel greyscale, over half of each file JPEG tables); (2) H.265 \emph{inside} the stream, encoded frame by frame only when the link is free so it never queues, records first, positions next (at $\le$10\,kbps through a 0.4$R$ token bucket), first key frame coarse (x265 \texttt{ipratio}=0.3) and, at $\le$10\,kbps, sent before the positions until it arrives; (3) an object-based ground in the spirit of MPEG-4 sprite coding (Smolic, Sikora and Ohm, IEEE TCSVT 9(8), 1999): a background plate with the objects inpainted every second plus one appearance crop per object, moved along the live positions. It was the most visible ground at 10\,kbps but lost to video on detector visibility and PSNR at 100\,kbps and 1\,Mbps (intra-only plates reuse nothing from the previous plate), and in dense scenes the crops starved the plates, so we dropped it; (4) the layer at 1\,Mbps to 10\,kbps, which keeps (2) and lets the video run up to 30\,fps; (5)--(8) the 1\,kbps chain below.

\paragraph{The final layer.} \emph{Ground, no extra bytes:} the newest decoded picture is warped to the camera of the newest position packet (the homography that packet carries); area it does not cover is filled from older pictures when their homography scores at least 20\,dB; when the picture is at least 0.2\,s older than the positions, every moving track (moved at least max(10\,px, half its box) within 1\,s against the scene) is cut out of the picture, its old place inpainted and the cut pasted at its position extrapolated from its last two updates (at most 1\,s). \emph{At 10 and 1\,kbps:} positions use a compact format (4-B header; a 16-B float16 homography only when the view moved more than 2\,px since the last one sent; 7\,B to introduce a track: slot, class, 12-bit centre, size; 3\,B for a move of at least 5\,px in 5-px steps; 1\,B when a track ends; still tracks omitted), coded against the last packet whose transmission started; a real encoder and decoder agree on all 16{,}701 delivered packets. A position packet is then 8--56\,B instead of a median 90\,B. The ground draws a hollow class-coloured box for each live track, predicts the camera's motion from the last two packets (at most 2\,s ahead) and applies it to picture and boxes; the moving-object transfer is off. 

\begin{table}[!t]
\centering\small
\setlength{\tabcolsep}{3pt}
\resizebox{\linewidth}{!}{%
\begin{tabular}{@{}lcccccc@{}}
\toprule
 & \multicolumn{3}{c}{ours} & \multicolumn{3}{c}{low-latency H.265 alone}\\
\cmidrule(lr){2-4}\cmidrule(l){5-7}
link & visible / covered & update (s) & age (s), PSNR & visible & new frame (s) & age (s), PSNR\\
\midrule
1\,Mbps & 0.67 / -- & 0.033 & 0.07, 31.7 & 0.52 & 0.033 & 0.27, 27.3\\
100\,kbps & 0.49 / -- & 0.045 (0.038) & 0.07, 29.2 & 0.33 & 0.032 & 0.37, 25.6\\
10\,kbps & 0.02 / 0.67 & 0.125 (0.064) & 0.27, 23.7 & 0.02 & 0.51 & 0.50, 23.5\\
1\,kbps$^\dagger$ & 0.00 / 0.10 & 2.30 (0.64) & 2.8, 20.0 & 0.00 & 1.26 & 6.0, 18.4\\
\bottomrule
\end{tabular}}
\caption{The final picture layer against low-latency H.265 alone (development). Visible: annotated objects the frozen detector finds in the displayed picture (IoU\,$\ge$\,0.3). Covered: annotated objects covered by any box the ground shows, detector or our object boxes; it favours our stream by construction. Update: median interval between new pictures (in brackets: between any display changes, including positions). Age: median picture age; PSNR to the current frame (dB). $^\dagger$The final 1\,kbps layer (8) on the 23 clips not used to choose its settings; at 10\,kbps the detector alone finds 0.018 of the annotated objects in our picture and 0.022 in video's; at 1\,kbps the colour base is on screen after 1.2\,s (video's first frame 1.95\,s); the 320-px picture in 21 of the 23 clips, median 20.7\,s; moving objects shown as their own pixels at a live position 5.6\% of the time (video's frames under 2\,s old: 6.3\%); boxes on their tracked object 65\% of the time at the frame the ground holds, 25\% at the current frame.}
\end{table}

\paragraph{The 1\,kbps chain (post-freeze, development only; each step declared with its measures before it ran, selection on the 24 odd-numbered clips, report on the other 23).} (5) Four picture settings (160$\times$90 CRF 40; 320$\times$180 CRF 40 and 44; 240$\times$136 CRF 42) with the uncovered area filled from the picture border: 160$\times$90 won, larger pictures arrive later and older. Then a two-regime scene: while the camera is slow (corner motion below 60\,px over 2\,s, hysteresis at 100\,px, a regime kept for 2\,s) a constant-QP stream at the video's own size with the QP chosen by a key-frame byte budget, a 160-px stream while it moves, the ground keeping the widest picture while the motion it infers from the positions is small; on the report half it beat video on 5 of 7 measures (objects covered 0.195 vs.\ 0, display every 0.47 vs.\ 1.26\,s, picture 3.1 vs.\ 6.0\,s old, 20.1 vs.\ 18.4\,dB, sharpness 16 vs.\ 5.5) and lost the picture interval (1.70 vs.\ 1.26\,s) and the first picture (3.8 vs.\ 1.95\,s). (6) The moving objects first as real pixels: an H.265 canvas of 24-px tiles, one per moving track, at most two refreshed per frame, each frame carrying its tracks' positions so it needs no prior packet. It showed moving objects as their own pixels at a live position 5.5\% of the time (video's frames under 2\,s old: 6.3\%) and moved the display every 0.37\,s, but paid with the scene: the picture became as old as video's (6.9 vs.\ 6.0\,s). (7) \emph{Positions:} the camera update as four corner offsets in 4-px steps (8\,B; 16\,B beyond $\pm$508\,px; a slot-keeping reset when a hop's homography scores below 20\,dB, because the compact update alone, skipped once, never advanced the ground's reference frame and every later box was wrong), 3-B size entries, introduction after 15 frames for moving and 30 for still tracks, a new track that continues one unseen for at most 1\,s (its box carried by its own velocity) keeps that track's identity, a track's end sent after 3--5 frames unseen and at once in a packet of its own, packets started at most every 0.5\,s (1\,s while the camera is slow) and encoded at the frame current when they start, the same 0.4$R$ share of the link in both regimes (the throttled variant, declared beside it, tied at 7 of 8 measures against video and lost the PSNR tie-break by 0.02\,dB). \emph{Scene:} a colour base (64$\times$36 or, while the camera moves, 32$\times$18 pixels; 100--200\,B) first and again when the camera has moved more than 100\,px since the newest scene picture or, while moving, every 2\,s; when the camera is slow a 320-px intra frame (the finest QP within 1.8\,kB) and then P-frame refinements against the previous layer, whose reference both ends recompute from the same decoded picture (deterministic x265, one frame thread; verified bit-identical). A ladder of P-frames climbing from the base was tried first and dropped: for the same bytes it is 40\% less sharp than one intra frame (a P-frame against a blurry reference skips blocks and keeps the blur). \emph{Scheduling:} records, then a due base, then the first object frame, then positions, then the scene until the intra frame is in, then object frames and refinements by deficit round robin (objects 1/5 while slow, 1/4 while moving; at most one object frame per 5\,s while moving). \emph{Measures added:} box precision at the frame the ground holds and at the current frame (IoU\,$\ge$\,0.3 against the onboard tracks), ghost share (boxes of tracks the onboard no longer reports), recall; the sharpness measure is taken on the picture before boxes are drawn (three intermediate runs had it inflated by the box outlines, which is recorded). What still separates the boxes from the current frame is the link's latency: packets arrive 0.6--1.5\,s apart, objects and camera move meanwhile, and a constant-velocity camera prediction made it worse on the moving-camera clip. (8) The final layer: as (7), but the 320-px intra frame (the finest QP within 1.2\,kB) is created as soon as the base is in, in either camera regime, is never dropped on a regime switch, and goes ahead of object frames and positions (apart from a keep-alive packet every 1.5--2\,s) until it arrives, the base being refreshed every 4\,s instead of 2\,s meanwhile. With (7) the picture reached only 9 of the 47 clips: most clips have a moving camera, and every attempt was dropped when the camera sped up. With (8) it reaches 45 of 47 (median 20.6\,s; the two without it are 7.5 and 12\,s long); a variant with a 1.8-kB frame reached 43 (median 22.0\,s) and tied on the selection half. The price is paid by the object layer and the positions while the frame is on the wire: moving objects shown as their own pixels 5.6\% instead of 8.2\% of the time, the display every 0.64 instead of 0.50\,s, 0.10 instead of 0.14 of the objects covered, and more of the frame left uncovered under a moving camera while the base refresh is slowed.

\section{Demonstration video}
A 133-second screen recording (ancillary file \texttt{demo.mp4}; no sound) shows three development scenes. Left: the onboard view with tracks, an encoding panel (our stream's bytes by kind on each link, the H.265 encoders' output per link, onboard compute per frame) and a data panel (bytes received, events opened and closed with their delay, the picture each ground holds, the audit trail of open, closed and withheld records). Right: our stream with the picture layer, plain H.265 and low-latency H.265 over the same four links (1\,Mbps to 1\,kbps), every ground decoding the actual bytes, and a 9-byte in-flight card update.

\end{document}